\documentclass[conference]{IEEEtran}
\IEEEoverridecommandlockouts

\usepackage{cite}
\usepackage{amsmath,amssymb,amsfonts}
\usepackage{algorithmic}
\usepackage{graphicx}
\usepackage{textcomp}
\usepackage{xcolor}
\usepackage{subfig}
\usepackage{gensymb}

\def\BibTeX{{\rm B\kern-.05em{\sc i\kern-.025em b}\kern-.08em
    T\kern-.1667em\lower.7ex\hbox{E}\kern-.125emX}}

\begin{document}
\title{Vision-Based Target Localization \\ for a Flapping-Wing Aerial Vehicle}

\author{\IEEEauthorblockN{1\textsuperscript{st} Xinghao Dong}
\IEEEauthorblockA{\textit{School of Automation and Electrical Engineering,}\\
	\textit{and Institute of Artificial Intelligence,} \\
	\textit{University of Science and Technology Beijing}\\
	Beijing 100083, China \\
	Email: g20208679@xs.ustb.edu.cn}
\and
\IEEEauthorblockN{2\textsuperscript{nd} Qiang Fu}
\IEEEauthorblockA{\textit{School of Automation and Electrical Engineering,}\\
	\textit{and Institute of Artificial Intelligence,} \\
	\textit{University of Science and Technology Beijing}\\
	Beijing 100083, China \\
	Email: fuqiang@ustb.edu.cn}
\and
\IEEEauthorblockN{3\textsuperscript{rd} Chunhua Zhang}
	\IEEEauthorblockA{\textit{Automation Research Institute Co.} \\
	\textit{Ltd. of China South Industries Group Corporation}\\
	Mianyang 621000, China \\
	Email: swaizhang@sina.com}
\and
\IEEEauthorblockN{4\textsuperscript{th} Wei He}
	\IEEEauthorblockA{\textit{School of Automation and Electrical Engineering,}\\
	\textit{and Institute of Artificial Intelligence,} \\
	\textit{University of Science and Technology Beijing}\\
	Beijing 100083, China \\
	Email: weihe@ieee.org}
\thanks{This work was supported by the National Key Research and Development Program of China under Grant 2019YFB1703603, the National Natural Science Foundation of China under Grant 61803025 and Grant 62073031, the Interdisciplinary Research Project for Young Teachers of USTB (Fundamental Research Funds for the Central Universities) under Grant FRF-IDRY-19-010 and the Fundamental Research Funds for the China Central Universities of USTB under Grant FRF-TP-19-001C2. Corresponding author is W. He.}
}
\maketitle
\begin{abstract}
The flapping-wing aerial vehicle (FWAV) is a new type of flying robot that mimics the flight mode of birds and insects. 
However, FWAVs have their special characteristics of less load capacity and short endurance time, so that most existing systems of ground target localization are not suitable for them.
In this paper, a vision-based target localization algorithm is proposed for FWAVs based on a generic camera model.
Since sensors exist measurement error and the camera exists jitter and motion blur during flight, Gaussian noises are introduced in the simulation experiment, and then a first-order low-pass filter is used to stabilize the localization values.
Moreover, in order to verify the feasibility and accuracy of the target localization algorithm, we design a set of simulation experiments where various noises are added.
From the simulation results, it is found that the target localization algorithm has a good performance.
\end{abstract}
\begin{IEEEkeywords}
FWAV, target localization, positioning algorithm, first-order low-pass filter
\end{IEEEkeywords}
\section{Introduction}
The flapping-wing aerial vehicle (FWAV) is a new type of flying robot that mimics the flight mode of birds and insects during flight \cite{b1}\cite{b2}\cite{b3}.
Compared with the conventional fixed-wing and rotary-wing aerial vehicles, the FWAVs are smaller in size with better flexibility during their flight, and they also hold great flight efficiency and a good biological concealment \cite{b4}\cite{b5}\cite{b6}.
These characteristics give them more advantages in the field of military reconnaissance, environmental monitoring, and disaster rescue \cite{b7}\cite{b8}\cite{b88}.

So far, there have been a lot of efforts in the research of FWAVs in recent years and many fruitful outcomes have also emerged.
Alireza et al. \cite{b9} in University of Illinois at Urbana-Champaign (UIUC) created a fully self-contained, autonomously flying robot with a mass of 93 g, called Bat Bot (B2), to mimic such morphological properties of bat wings. 
Additionally, Aero Vironment in the U.S. \cite{b10} designed a FWAV named hummingbird with a mass of 19 g and a wingspan of 16.5 cm. 
Moreover, Yang et al. from Northwestern Polytechnical University also created a FWAV named Dove. It has a mass of 220 g, a wingspan of 50 cm, and the ability to operate autonomously. It can fly lasting half an hour, and transmit live stabilized colorful videos to a ground station over 4 km away \cite{b11}.

In addition, a new dynamic analysis and engineering design method for flapping-wing flying robots is proposed in the background of engineering application \cite{b12}.
Researchers from the Delft University of Technology designed a micro FWAV named DelFly II \cite{b13}. It equips a binocular vision system, which consists of two synchronous CMOS cameras with a total resolution of 720$\times$240 pixels. Using this vision system, DelFly II can carry out some simple obstacle detection tasks.
However, its measurement distance is only 4 m and the error even reaches 30 cm, so the visual perception system is not suitable for complex tasks.
Moreover,
Baek et al. from the University of California, Berkeley, demonstrated a 13-gram autonomously controlled ornithopter by using a small X-Wing \cite{b14}. 
With a wingspan of 28 cm, it can land within a radius of 0.5 m from the target, and the success rate reaches more than 85\%. 
Although the function of flying to the designated target independently has been realized, this system is not suitable for a FWAV with big wingspan.

Generally, the ground target localization system for a FWAV has not received much attention, but a great development has been achieved for other aircraft platforms instead \cite{b15}.
As for aircraft-based ground target localization methods, there are three types at present: target location methods based on attitude measurement and laser ranging; target localization methods based on collinear conformation principle; target localization methods based on Digital Elevation Model (DEM) \cite{b17}.
However, FWAVs have their own characteristics of less load capacity, short endurance time and motion 
jitter during flight, so most existing systems are not suitable for FWAVs.

In this paper, we concentrate on designing a ground target localization system for FWAVs.
First, we introduce the generic camera model that we have chosen and explain the reason.
Based on the generic camera model, the ground target localization algorithm is given next. 
In addition, considering the measurement errors of sensors and motion jitter during flight, when we then do the simulation experiments, we introduce Gaussian noises to the variables of height and camera angle between the camera and the target.
Moreover, to suppress the influence of noises on the localization system, localization values are filtered by a first-order low-pass filter.

The paper is organized as follows.
Five coordination systems and the generic camera model will be introduced firstly in Section II.
In Section III, the target localization algorithm will be presented.
And in Section IV, various simulation experiments are carried out. 
Experimental results are discussed and concluded in Section V.
\section{Preliminaries}
\subsection{Five coordinate systems}
\begin{figure}
	\centering
	\includegraphics[width=0.9\linewidth, height=0.28\textheight]{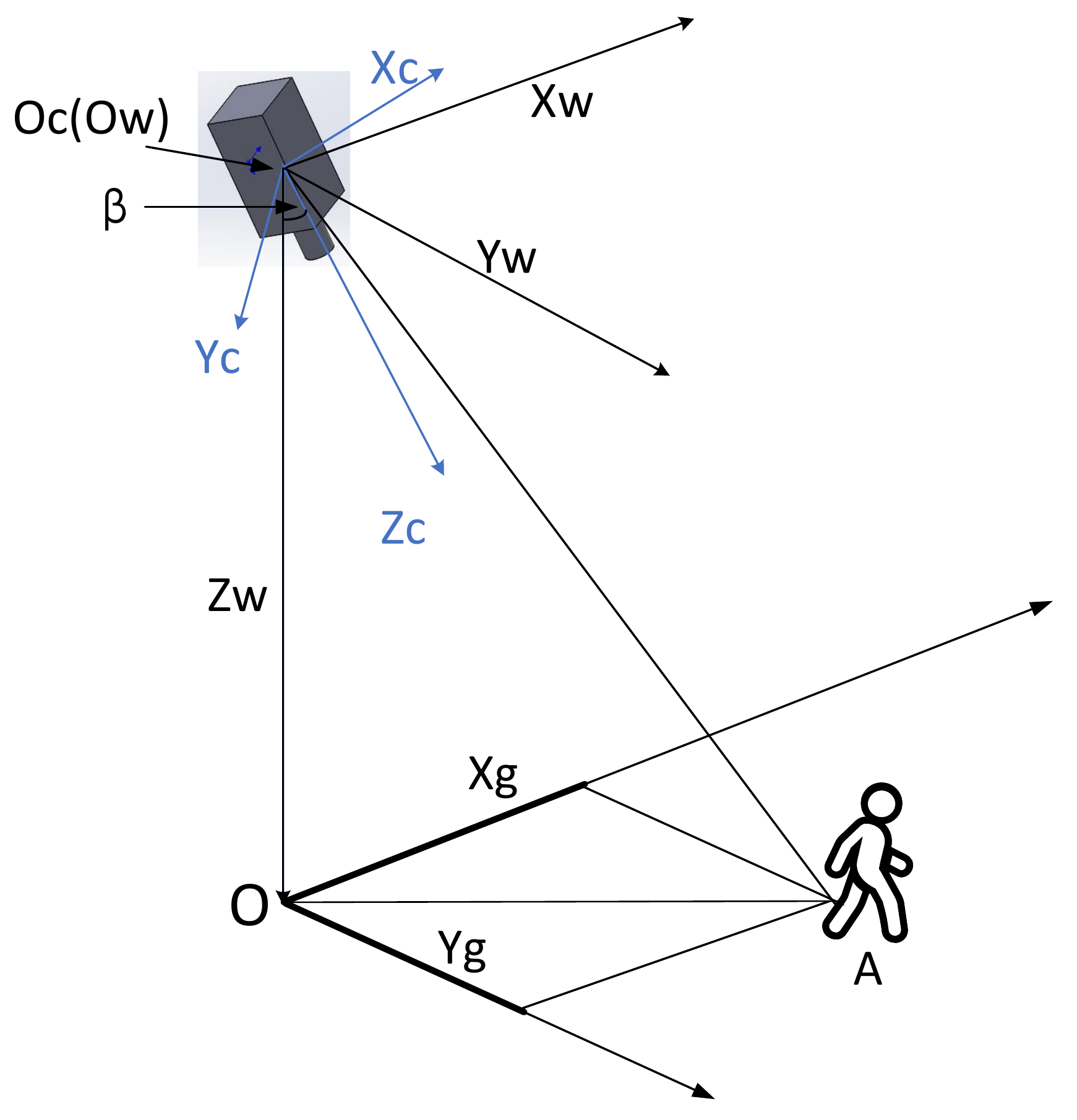}
	\caption{World coordinate system and camera coordinate system}
	\label{Ground target localization diagram}
\end{figure}
As shown in Fig.\ref{Ground target localization diagram}, there are three coordinate systems. First, assuming a man (point A) as the target, we define that the ground coordinate system is $X_{\mathrm{g}}$$O$$Y_{\mathrm{g}}$, and the $O$$X_{\mathrm{g}}$ and $O$$Y_{\mathrm{g}}$ axes point to the east and north respectively.
Moreover, supposing point $O_{\mathrm{c}}$ ($O_{\mathrm{w}}$) is the centroid of the camera, the world coordinate system is $O_{\mathrm{w}}$-$X_{\mathrm{w}}$$Y_{\mathrm{w}}$$Z_{\mathrm{w}}$. The point $O$ is the perpendicularly vertical point of $O_{\mathrm{c}}$ on the ground coordinate system, and the $O$$X_{\mathrm{g}}$ and $O$$Y_{\mathrm{g}}$ axes are parallel to the $O_{\mathrm{w}}$$X_{\mathrm{w}}$ and $O_{\mathrm{w}}$$Y_{\mathrm{w}}$ axes respectively.
Additionally, the camera coordinate system $O_{\mathrm{c}}$-$X_{\mathrm{c}}$$Y_{\mathrm{c}}$$Z_{\mathrm{c}}$ is formed as follows: take the camera centroid $O_{\mathrm{c}}$ as origin, take the camera optical axis as the $O_{\mathrm{c}}Z_{\mathrm{c}}$ axis, and the $X_{\mathrm{c}}$, $Y_{\mathrm{c}}$, $Z_{\mathrm{c}}$ axes are perpendicular to each other as shown in Fig.\ref{Ground target localization diagram}.
The angle between $Z_{\mathrm{w}}$ and $Z_{\mathrm{c}}$ is $\beta$.
\begin{figure}
	\centering
	\includegraphics[width=0.9\linewidth, height=0.17\textheight]{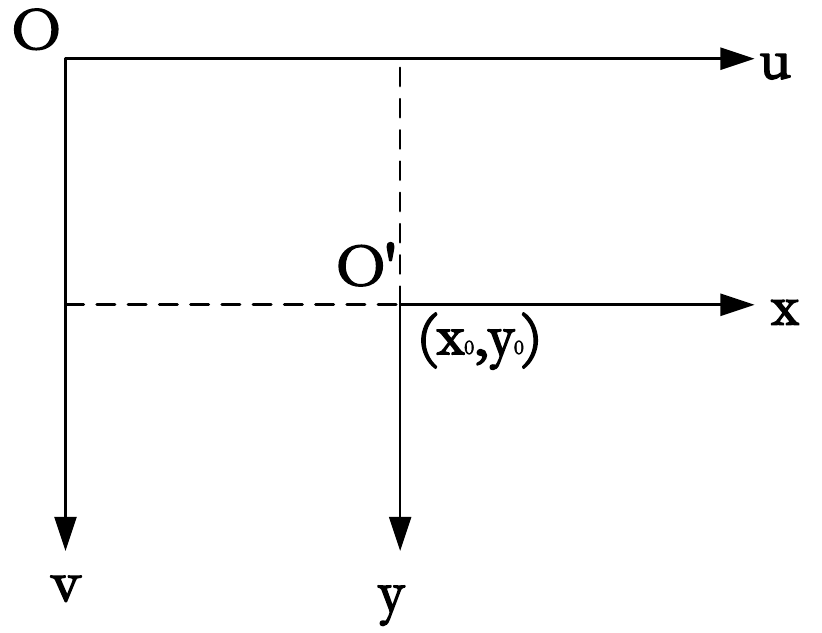}
	\caption{Image coordinate systems}
	\label{Figure coordinate system and pixel coordinate system}
\end{figure}

Besides, as shown in Fig.\ref{Figure coordinate system and pixel coordinate system}, there are two image coordinate systems
called image coordinate system I (${xO'y}$) and image coordinate system II (${uOv}$).
Both of them are on the imaging plane, but their origin and measurement unit are different. The origin ($O'$) of the image coordinate system I is the midpoint of the imaging plane, while the origin ($O$) of the image coordinate system II is on the top left corner. Moreover, the unit of image coordinate system I is mm, which belongs to its physical unit, while the unit of image coordinate system II is pixel.
In addition, the conversion formula that from image coordinate system I to image coordinate system II is
\begin{equation}
	\left[\begin{array}{l}
		x \\
		y
	\end{array}\right]=
	\left[\begin{array}{cc}
		m_{\mathrm{x}} & 0 \\
		0 & m_{\mathrm{y}}
	\end{array}\right]
	\left[\begin{array}{l}
		u_{\mathrm{}} \\
		v_{\mathrm{}}
	\end{array}\right]+
	\left[\begin{array}{l}
		x_{\mathrm{0}} \\
		y_{\mathrm{0}}
	\end{array}\right]
\end{equation}
where $(x_{\mathrm{0}},y_{\mathrm{0}})^T$ is the principal point and $m_{\mathrm{x}}$, $m_{\mathrm{y}}$ give the number of pixels per unit distance (mm) in horizontal and vertical directions respectively \cite{b18}.
\subsection{Generic camera model}
\begin{figure}
	\centering
	\includegraphics[width=0.72\linewidth, height=0.25\textheight]{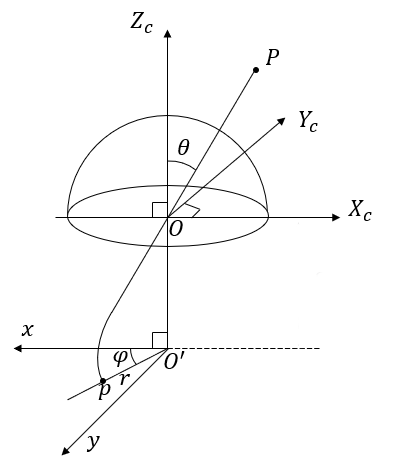}
	\caption{Fish-eye lenses model.}
	\label{Fish-eye lenses model}
\end{figure}

In order to increase the field of view of aerial photography based on FWAVs, a wide-angle lenses camera is usually chosen to be used \cite{b19}.
However, the traditional pinhole image model is not suitable for wide-angle cameras, so we adopted a generic fish-eye lenses camera model as shown in Fig.\ref{Fish-eye lenses model}.

Point $O$ is the center of the camera lenses, $O'$ is the image center, $\theta$ is the angle of incidence of projection, and $r$ is the distance from the projection point $p$ to the image center $O'$. 
The plane $xOy$ is the image surface, point $p$ is the image point of the space point $P$, and $\varphi$ is the angle between $O'p$ and $x$ axis.
The approximate projection model is as follows
\begin{equation}
	r(\theta)=k_{\mathrm{1}}\theta+k_{\mathrm{2}}\theta^3+k_{\mathrm{3}}\theta^5+k_{\mathrm{4}}\theta^7+k_{\mathrm{5}}\theta^9
\end{equation}
In the above equation, $k_{\mathrm{1}}$, $k_{\mathrm{2}}$, $k_{\mathrm{3}}$, $k_{\mathrm{4}}$, $k_{\mathrm{5}}$ represent 5 parameters which can be determined by the calibration procedure.

\begin{figure}
	\centering
	\includegraphics[width=0.95\linewidth, height=0.05\textheight]{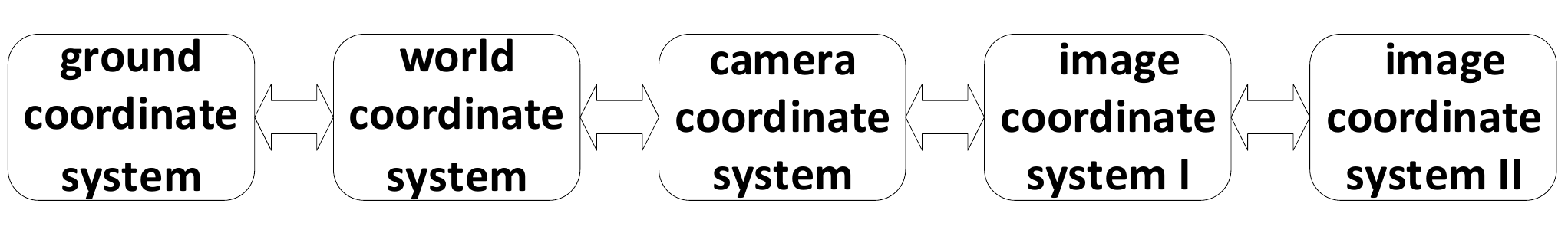}
	\caption{Coordinate system transformation}
	\label{Projection coordinate transformation}
\end{figure}
Based on the formula (2), the coordinate system transformation process is shown in Fig.\ref{Projection coordinate transformation}.
And the projection process is to calculate the coordinate of the target in image coordinate system II for giving its ground coordinate. 

Suppose that the coordinate of the ground target in the ground coordinate system is ${A_g} = {(x_{\mathrm{A}},y_{\mathrm{A}})^T}$, the height of the camera is $h$, and the angle between $Z_{\mathrm{w}}$ and $Z_{\mathrm{c}}$ is $\beta$. Then the point A in the world coordinate system is ${A_w} = ({{x_{\mathrm{A}}},{y_{\mathrm{A}}},{h}})^T$. The point in the world coordinate system is transformed into the camera coordinate system by rigid transformation matrix $R$
\begin{equation}
	R=
	\left[\begin{array}{ccc}
		\cos\beta & 0 & -\sin\beta \\
		0 & 1 & 0 \\
		\sin\beta & 0 & \cos\beta
	\end{array}\right]
\end{equation}
Then the coordinate of the ground target point in the camera coordinate system is $A_{\mathrm{c}}=\left(X_{\mathrm{c}}, Y_{\mathrm{c}}, Z_{\mathrm{c}}\right)^T$
\begin{equation}
	A_{\mathrm{c}}=RA_{\mathrm{w}}
\end{equation}
Then we can get the incidence angle $\theta$, and the angle $\varphi$ (as shown in Fig.\ref{Fish-eye lenses model}) as follows
\begin{equation}
	\theta=\arccos(Z_{\mathrm{c}}/\sqrt{X_{\mathrm{c}}^{2}+Y_{\mathrm{c}}^{2}+z_{\mathrm{c}}^{2}})
\end{equation}
\begin{equation}
	\varphi=\arccos(X_{\mathrm{c}}/\sqrt{X_{\mathrm{c}}^{2}+Y_{\mathrm{c}}^{2}})
\end{equation}
By combining (2), we can get the distance $r$.
And then we can get the coordinate $(x,y)^T$ in the image coordinate system I 
\begin{equation}
	\left[\begin{array}{l}
		x \\
		y
	\end{array}\right]=r\left[\begin{array}{l}
		cos\varphi \\
		sin\varphi
	\end{array}\right]
\end{equation}
And then we can get the coordinate $(u,v)^T$ in the image coordinate system II
\begin{equation}
	u=1/m_{\mathrm{x}}\cdot(x-x_{\mathrm{0}})
\end{equation}
\begin{equation}
	v=1/m_{\mathrm{y}}\cdot(y-y_{\mathrm{0}})
\end{equation}
\section{GROUND TARGET POSITIONING ALGORITHM}
The target localization algorithm aims to calculate the coordinate of the ground target in the ground coordinate system based on knowing its image coordinate.
\begin{figure}
	\centering
	\includegraphics[width=0.8\linewidth, height=0.13\textheight]{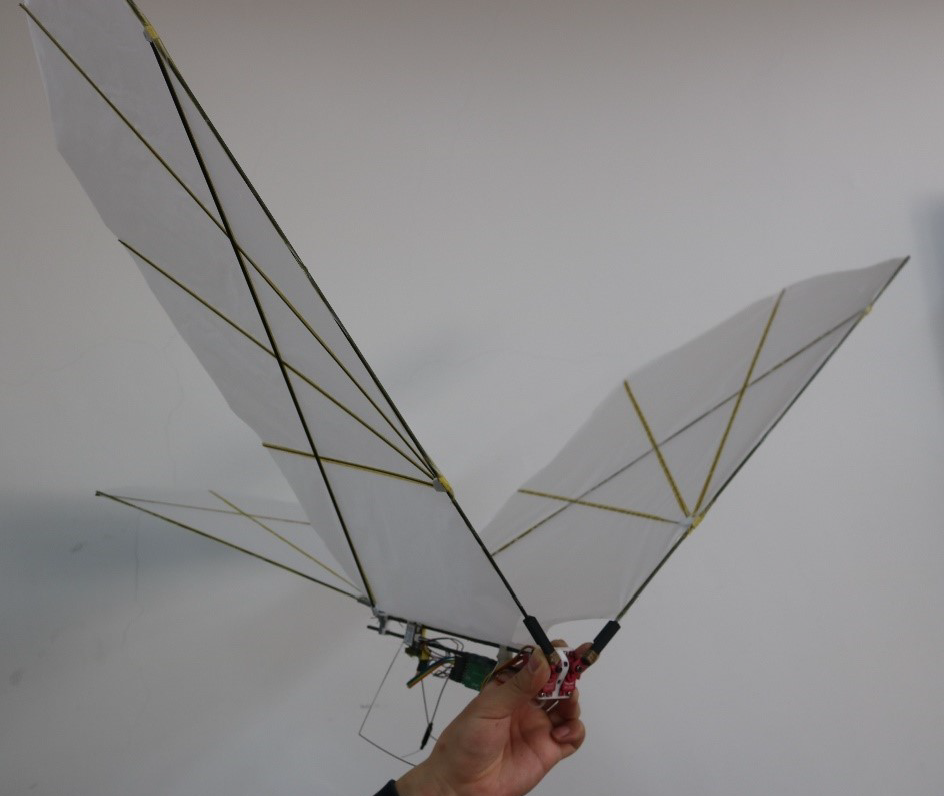}
	\caption{USTBird}
	\label{USTBird}
\end{figure}

In addition, Fig.\ref{USTBird} shows one of the FWAVs named USTBird.
Note that sensors exist measurement error and the camera also exists jitter and
motion blur during flight. When we then calculate the coordinate of the target in the ground coordinate system, we suppose that the angle between $Z_{\mathrm{w}}$ and $Z_{\mathrm{c}}$ is $\beta'$, the height of the camera is $h'$, and the coordinate in image coordinate system II is $(u',v')^T$.

Given the coordinate $(\mathrm{u'},\mathrm{v'})^T$ as mentioned, combining (1), we can get the coordinate $(x',y')$ in the image coordinate system I.
Thus, the distance $pO'$ ($r_{\mathrm{c}}$) and the angle between the $pO'$ and the $x$ axis ($\varphi_{\mathrm{c}}$) are
\begin{equation}
	r_{\mathrm{c}}=\sqrt{x'^{2}+y'^{2}}
\end{equation}
\begin{equation}
	\varphi_{\mathrm{c}}=\arctan(y'/x')
\end{equation}
And then combining (2), we can get the incidence angle $\theta_{c}$ by solving the nine-degree polynomial given the internal parameters of the camera.
Additionally, given the incident angle $\theta_{c}$ and the angle $\varphi_{\mathrm{c}}$, the unit vector of incident angle in camera coordinate system ${V_c} = {(x_{vc},y_{vc},z_{vc})^T}$ is
\begin{equation}
	x_{\mathrm{vc}}=\cos\varphi_{\mathrm{c}}\cdot\sin\theta_{\mathrm{c}}
\end{equation}
\begin{equation}
	y_{\mathrm{vc}}=\sin\varphi_{\mathrm{c}}\cdot\sin\theta_{\mathrm{c}}
\end{equation}
\begin{equation}
	z_{\mathrm{vc}}=\cos\theta_{\mathrm{c}}
\end{equation}
Then according to (3), the rotation matrix $R'$ from camera coordinate system to world coordinate system is
\begin{equation}
	R'=\left[\begin{array}{ccc}
		\cos\beta' & 0 & -\sin\beta' \\
		0 & 1 & 0 \\
		\sin\beta' & 0 & \cos\beta'
	\end{array}\right]
\end{equation}
In addition, according to (4), we can then get the unit vector of incident angle in the world coordinate system ${V_w} = {(x_{vw},y_{vw},z_{vw})^T}$
\begin{equation}
	V_{\mathrm{w}}=R'^{-1} \cdot V_{\mathrm{c}}
\end{equation}
Then, combining (5) and (6), based on the the unit vector, the angles $\theta_{w}$ and $\varphi_{w}$ are
\begin{equation}
	{\theta_w} = \arccos ({z_{vw}}/\sqrt {{x_{vw}}^2 + {y_{vw}}^2 + {z_{vw}}^2} )
\end{equation}
\begin{equation}
	{\varphi_w} = \arccos ({x_{vw}}/\sqrt {{x_{vw}}^2 + {y_{vw}}^2} )
\end{equation}
Based on all the above formulas, we can finally obtain the coordinate of the target in the ground coordinate system given the height of the camera $h'$.
And the coordinate $({X_{g}},{Y_{g}})^T$ is
\begin{equation}
	{X_{g}} = h' \cdot \tan {\theta _w} \cdot \cos {\varphi _w}
\end{equation}
\begin{equation}
	{Y_{g}} = h' \cdot \tan {\theta _w} \cdot \sin {\varphi _w}
\end{equation}
\section{SIMULATION EXPERIMENT}
\subsection{Simulation devices and configurations}
In order to verify the feasibility and accuracy of the localization algorithm, we designed a set of comparative simulation experiments. 
The simulation code is written and run by python 3.7, and the experimental platform is a laptop with a 2.9 GHz CPU and 16 G memory, and the operating system is windows 10.

Moreover, according to our real cameras, we set parameters of the generic camera model with $u_{\mathrm{0}}=320$, $v_{\mathrm{0}}=240$, $m_{\mathrm{u}}=188$, $m_{\mathrm{v}}=188$, and the maximum camera angle is 150\textdegree.
After calibration, we set the parameters in the projection model with $k_{\mathrm{1}}=3.55$, $k_{\mathrm{2}}=0.03$, $k_{\mathrm{3}}=k_{\mathrm{4}}=k_{\mathrm{5}}=0$.
%
%
\subsection{Simulation with noises }
The errors are caused by the measurement of sensors and signal transmission, which include the angle error approximately 0.1\degree, the height error introduced by the barometer is 0.5 m, and the image detection error is almost 4 pixels.
Moreover, the flight mode of flapping-wing aerial vehicles will cause periodic jitter in the fuselage during flight, leading to jitter and motion blur to the camera.
So we then introduced Gaussian noises in the simulation experiment.
\begin{figure}
	\centering
	\includegraphics[width=0.97\linewidth, height=0.225\textheight]{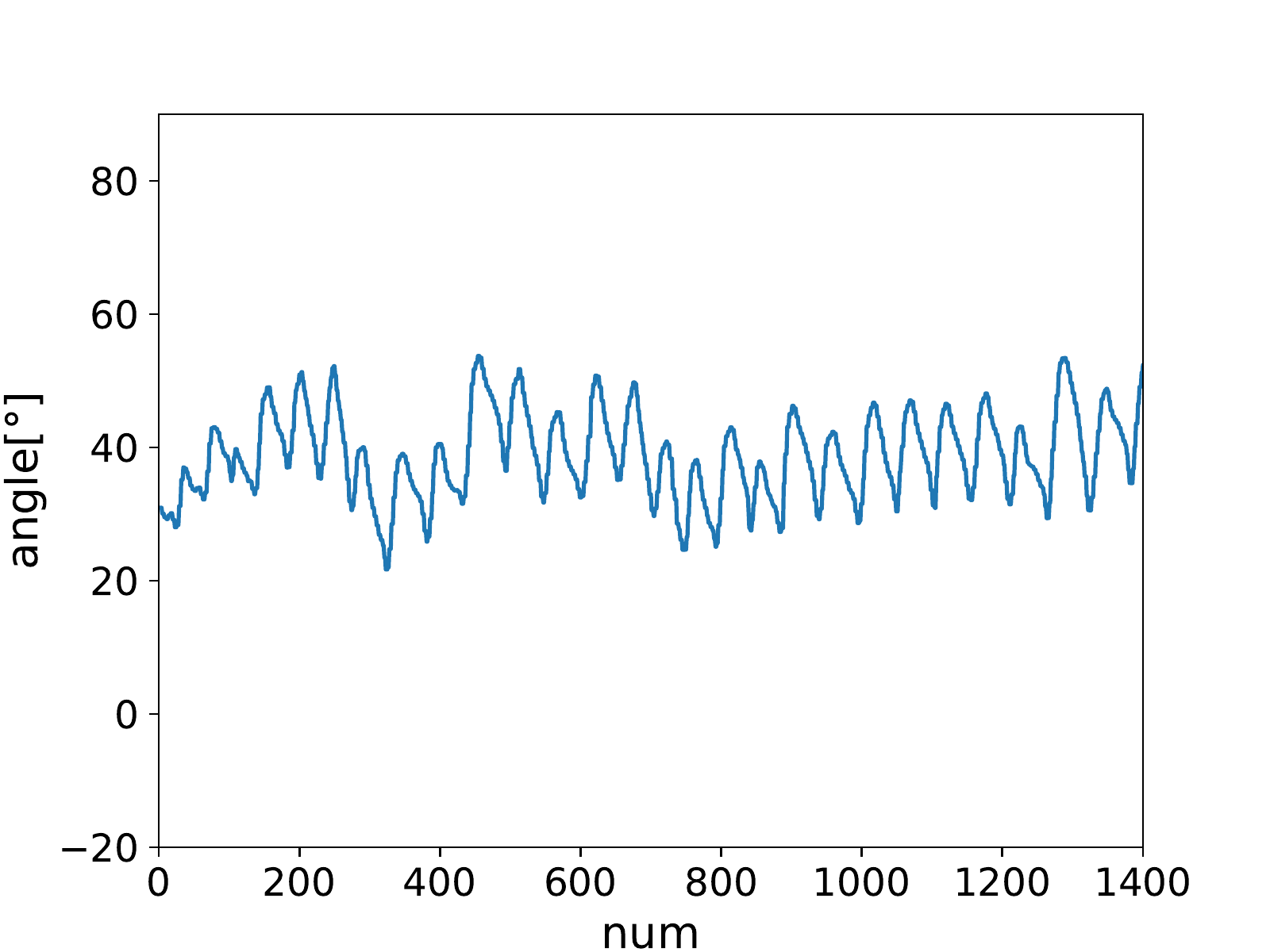}
	\caption{Pitch attitude angle of USTBird}
	\label{Pitch attitude angle of USTBird}
\end{figure} 

Fig.\ref{Pitch attitude angle of USTBird} shows the attitude pitch angle of USTBird during the flight. The pitch of USTBird fluctuates periodically around 37\degree \ showing that the attack angle of USTBird is about 37\textdegree.
So we then introduced the Gaussian noise in angle part with the mean value of 37\degree \ 
and with the standard deviation of 0.1\degree \ (the respective error level of the sensor as mentioned before). While in the height part, the Gaussian noise is introduced with the mean value of 0 and the standard deviation of 0.5 m (the respective error level of the sensor as mentioned before).

In doing the simulation experiment, we set a fixed point $(x,y,height)^T$=$(4,3,10)^T$ as our target coordinate in the world coordinate system, and then calculate the coordinate by our localization algorithm many times.

Specifically, the X and Y coordinates are calculated 90 times respectively, and the simulation results are shown in Fig.\ref{X coordination} and Fig.\ref{Y coordination} with the dotted green lines.

As the noise existed in both part of measurement and motion jitter, both of the X and Y coordinate lines are not very smooth even with some shaking points in a very large attitude.
So we should then introduce a filter to stabilize the localization values.
\subsection{First-order low-pass filter}
The first-order low-pass filtering algorithm originated from the first-order low-pass filter circuit, which is used to eliminate superimposed high-frequency interference components \cite{b20}.
We could use a first-order low-pass filter to stabilize the localization values influenced by high-frequency noise of the measurement error and motion jitter.

The algorithm formula of the first-order low-pass filter is 
\begin{equation}
	u_{k} = \alpha x_{k-1} + (1 - \alpha )u_{k-1}
\end{equation}
where $x_{k-1}$ is the $(k-1)$th value of sampling, $u_{k-1}$ is the $(k-1)$th value of filter output, $u_{k}$ is the $k$th output value of the filter, and $\alpha$ is the filtering coefficient \cite{b21}.

\begin{figure}
	\centering
	\includegraphics[width=1.051\linewidth, height=0.2\textheight]{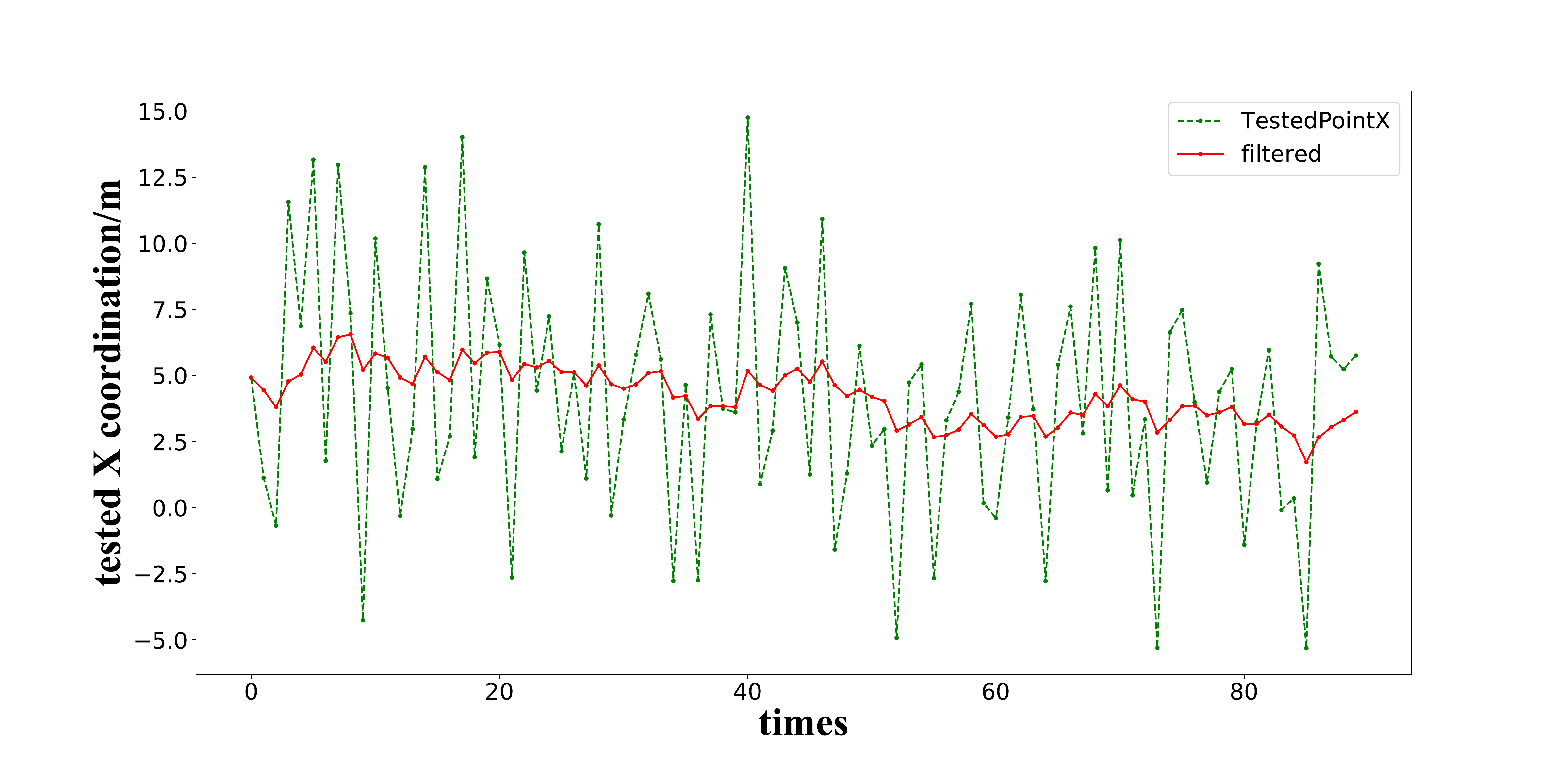}
	\caption{X coordination}
	\label{X coordination}
\end{figure}

\begin{figure}
	\centering
	\includegraphics[width=1.051\linewidth, height=0.2\textheight]{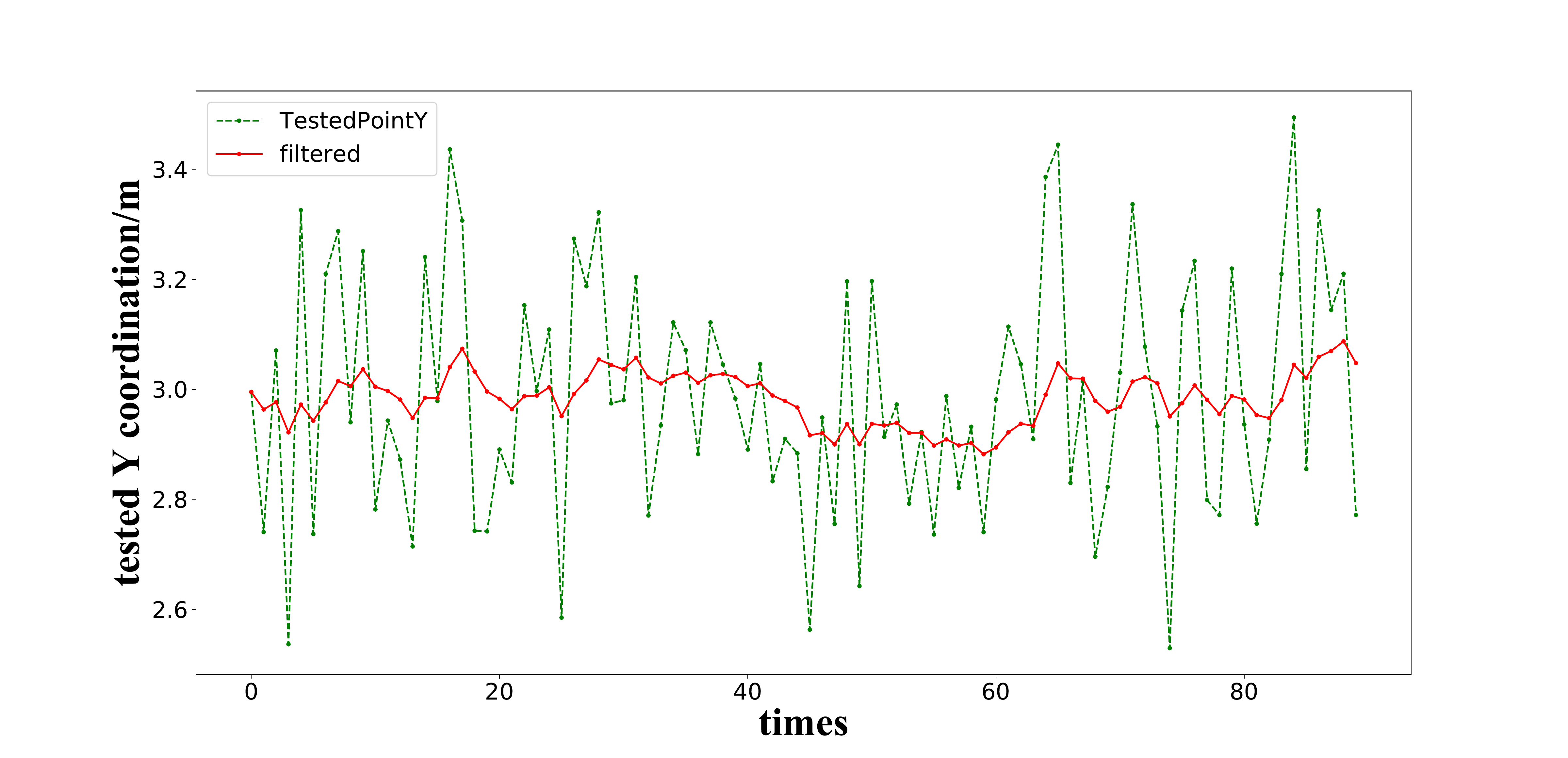}
	\caption{Y coordination}
	\label{Y coordination}
\end{figure}

After testifying the filtering coefficient $\alpha$ many times, we finally chose the filtering coefficient $\alpha$=0.125, which can receive a better filtering effect.
The filtered result is shown in Fig.\ref{X coordination} and Fig.\ref{Y coordination} with the red line, and the data analysis results are shown in Table I.


\begin{table}[]
	\renewcommand\arraystretch{2.0}
	\caption{Data analysis}
	\vspace{1.5pt}
	\centering
	\tabcolsep=2.5pt
	\begin{tabular}{ccccc}
		\hline
		&Mean (original) & Mean (filtering) & Var. (original) & Var. (filtering) \\
		\hline
		X / m  & 4.54& 4.11& 18.76& 1.24  \\
		
		Y / m  & 3.46& 3.17& 15.74& 2.51  \\
		\hline
	\end{tabular}
	\label{bs2}
\end{table}

As shown in Table I, the mean value is closer to the set point, and the variance is greatly reduced after filtering.
The simulation results show that the filter can eliminate the high-frequency interference and stabilize the localization value to our set real position value.
In short, from the simulation results, the localization algorithm and the filter we used have a good effect.
\section{CONCLUSION}
In this paper, we mainly developed a vision-based target localization algorithm for FWAVs.
The five coordinates and generic camera model are introduced firstly, and based on that camera model, a target localization algorithm is presented.
Moreover, we then carried out various simulation experiments. 
Considering measurement errors of sensors and motion fuselage jitter during flight, the simulation is processed by adding Gaussian noises to the variables of height and camera angle.
In addition, we used a first-order low-pass filter to stabilize the calculated localization value influenced by the noises.
The simulation results show that the target localization algorithm has a good effect, and the filter we used also makes a grea improvement.
In the future, we will evaluate the performance of the proposed ground target localization algorithm in real experiments.


\begin{thebibliography}{00}
	\bibitem{b1} W. He, X. X. Mu, L. Zhang, and Y. Zou, ``Modeling and trajectory tracking
	control for flapping-wing micro aerial vehicles,'' IEEE/CAA Journal of
	Automatica Sinica, vol. 8, no. 1, pp. 148--156, 2021.
	\bibitem{b2} Z. Yin, W. He, Y. Zou, X. X. Mu and C. Y. Sun, ``Efficient formation of flapping-wing aerial vehicles based on wild geese queue effect,'' Acta Automatica Sinica, vol. 46, pp. 1--13, 2020.
	\bibitem{b3} Q. Fu, X. Y. Chen, Z. L. Zheng, Q. Li, and W. He, ``Research progress
	on visual perception system of bionic flapping-wing aerial vehicles,''
	Chinese Journal of Engineering, vol. 41, no. 12, pp. 1512--1519, 2019.
	\bibitem{b4} M. A. Graule, P. Chirarattananon, S. B. Fuller, N. T. Jafferis, K. Y. Ma,
	M. Spenko, R. Kornbluh, and R. J. Wood, ``Perching and takeoff of
	a robotic insect on overhangs using switchable electrostatic adhesion,''
	Science, vol. 352, no. 6288, pp. 978--982, 2016.
	\bibitem{b5} F. Y. Hsiao, L. J. Yang, S. H. Lin, C. L. Chen, and J. F. Shen, ``Autopilots
	for ultra lightweight robotic birds: Automatic altitude control and system
	integration of a sub-10 g weight flapping-wing micro air vehicle,'' IEEE
	Control Systems Magazine, vol. 32, no. 5, pp. 35--48, 2012.
	\bibitem{b6} E. Pan, X. Liang, and W. Xu, ``Development of vision stabilizing system
	for a large-scale flapping-wing robotic bird,'' IEEE Sensors Journal,
	vol. 20, no. 99, pp. 8017--8028, 2020.
	\bibitem{b7} J. Pietruha, K. Sibilski, M. Lasek, and M. Zlocka, ``Analogies between
	rotary and flapping wings from control theory point of view,'' in AIAA
	Atmospheric Flight Mechanics Conference and Exhibit, 2001.
	\bibitem{b8} C. S. Yuan, Y. Z. Li, and J. Tan,
	``Investigation in flight control
	system of flapping-wing micro air vehicles,'' Computer Measurement
	and Control, vol. 19, no. 7, pp. 1527--1529, 2011.
	\bibitem{b88} Q. Fu, J. Wang, L. Gong, J. Y. Wang, ang W. He, ``Obstacle avoidance of flapping-wing air vehicles based on
	optical flow and fuzzy control,'' Transactions of Nanjing University of Aeronautics and Astronautics, vol. 38, no. 2, pp. 206--215, 2021.
	\bibitem{b9} A. Ramezani, X. Shi, S. J. Chung, and S. Hutchinson, ``Bat Bot (B2), a biologically inspired flying machine,''
	in IEEE International Conference on Robotics and Automation, 2016.
	\bibitem{b10} M. Keennon, K. Klingebiel, and H. Won, ``Development of the nano
	hummingbird: A tailless flapping wing micro air vehicle,'' in 50th AIAA
	Aerospace Sciences Meeting including the New Horizons Forum and
	Aerospace Exposition, 2012.
	\bibitem{b11} W. Yang, L. Wang, and B. Song, ``Dove: A biomimetic flapping-wing
	micro air vehicle,'' International Journal of Micro Air Vehicles, vol. 10,
	no. 1, pp. 70--84, 2018.
	\bibitem{b12} A. A. Paranjape, S. J. Chung, and J. Kim, ``Novel dihedral-based
	control of flapping-wing aircraft with application to perching,'' IEEE
	Transactions on Robotics, vol. 29, no. 5, pp. 1071--1084, 2013.
	\bibitem{b13} G. C. H. E. D. Croon, M. Perin, B. D. W. Remes, R. Ruijsink, and C. D.
	Wagter, The DelFly: Design, Aerodynamics, and Artificial Intelligence
	of a Flapping Wing Robot, Springer Publishing Company, Incorporated,
	2015.
	\bibitem{b14} S. S. Baek, F. L. G. Bermudez, and R. S. Fearing, ``Flight control for
	target seeking by 13 gram ornithopter,'' in 2011 IEEE/RSJ International
	Conference on Intelligent Robots and Systems, 2011.
	\bibitem{b15} Z. Zhang, B. Han, P. Li, F. Zhou, and W. Xu, ``Small unmanned aerial
	vehicle visual system for ground moving target positioning,'' in International
	Conference on Automatic Control and Artificial Intelligence,
	2012.
	\bibitem{b17} W. Han, J. Wang, N. Wang, G. Sun, and D. He, ``A method of ground
	target positioning by observing radio pulsars,'' Experimental Astronomy,
	vol. 49, no. 1, pp. 43--60, 2020.
	\bibitem{b18} J. Kannala and S. S. Brandt, ``A generic camera model and calibration
	method for conventional, wide-angle, and fish-eye lenses,'' IEEE Transactions
	on Pattern Analysis and Machine Intelligence, vol. 28, no. 8,
	pp. 1335--1340, 2006.
	\bibitem{b19} Q. Fu, Y. H. Yang, X. Y. Chen, and Y. L. Shang,
	``Vision-based obstacle avoidance for flapping-wing aerial vehicles,''
	Science China Information Sciences, vol. 63, no. 7, pp. 170208,
	2020.
	\bibitem{b20} Y. Wang, Z. Chen, M. Sun, and Q. Sun, ``Ladrc-smith controller design
	and parameters analysis for first-order inertial systems with large timedelay,''
	in 2018 IEEE 7th Data Driven Control and Learning Systems
	Conference (DDCLS), 2018.
	\bibitem{b21} Y. Wang, D. Y. Sun, and Z. G. Qi, ``Aging simulation method for EFI engine oxygen sensor based on first order inertial filter algorithm,'' Jilin
	Daxue Xuebao (Gongxueban) /Journal of Jilin University (Engineering
	and Technology Edition), vol. 47, no. 4, pp. 1040--1047, 2017.
\end{thebibliography}
\end{document}